# Dependent Hierarchical Normalized Random Measures for Dynamic Topic Modeling

**Changyou Chen**[1,3] CCHANGYOU@GMAIL.COM
[1]Research School of Computer Science, The Australian National University, Canberra, ACT, Australia

**Nan Ding**[2] DING10@PURDUE.EDU
[2]Department of Computer Science, Purdue University, USA

**Wray Buntine**[3,1] WRAY.BUNTINE@NICTA.COM.AU
[3]National ICT, Canberra, ACT, Australia

## Abstract

We develop dependent hierarchical normalized random measures and apply them to dynamic topic modeling. The dependency arises via *superposition*, *subsampling* and *point transition* on the underlying Poisson processes of these measures. The measures used include normalised generalised Gamma processes that demonstrate power law properties, unlike Dirichlet processes used previously in dynamic topic modeling. Inference for the model includes adapting a recently developed slice sampler to directly manipulate the underlying Poisson process. Experiments performed on news, blogs, academic and Twitter collections demonstrate the technique gives superior perplexity over a number of previous models.

## 1. Introduction

Dirichlet processes and their variants are popular in recent years, with applications found in diverse discrete domains such as topic modeling (Teh et al., 2006), *n*-gram modeling (Teh, 2006), clustering (Socher et al., 2011), and image modeling (Li et al., 2011). These models take as input a base distribution and produce as output another distribution which is somewhat similar. Moreover, they can be used hierarchically. Together this makes them ideal for modeling structured data such as text and images.

When modeling dynamic data or data from multiple sources, dependent nonparametric Bayesian models (MacEachern, 1999) are needed in order to harness related or previous information. Among these models, the hierarchical Dirichlet process (HDP) (Teh et al., 2006) is the most popular one. However, a basic assumption underlying the HDP is the full exchangeability of the sample path, which is often violated in practice, *e.g.*, we could assume the content of ICML depends on previous years' so order is important.

To overcome the full exchangeability limitation, several dependent Dirichlet process models have been proposed, for example, the dynamic HDP (Ren et al., 2008), the evolutionary HDP (Zhang et al., 2010), and the recurrent Chinese Restaurant process (Ahmed & Xing, 2010). Dirichlet processes are used because of simplicity and conjugacy (James et al., 2006). These models are constructed by incorporating the previous DP's into the base distribution of the current DP. Dependent DPs have also been constructed using the underlying Poisson processes (Lin et al., 2010). However, recent research has shown that many real datasets have the power-law property, *e.g.*, in images (Sudderth & Jordan, 2008), in topic-word distributions (Teh, 2006) and in document topic (label) distributions (Rubin et al., 2011). This makes the Dirichlet process an improper tool for modeling these datasets.

Although there also exists some dependent nonparametric models with power-law phenomena, their dependencies are limited. For example, Bartlett et al. (2010) proposed a dependent hierarchical Pitman-Yor process that only allows deletion of atoms, while Sudderth & Jordan (2008) construct the dependent Pitman-Yor process by only allowing dependencies between atoms.

In this paper, we use a larger class of stochastic processes called normalized random measures with independent increments (NRM) (James et al., 2009). While this includes the Dirichlet process as a special case, some other versions of NRMs have the power-law property. This class of discrete random measures can also be constructed from Poisson processes. Given this, following (Lin et al., 2010), we analogically define *superposition*, *subsampling* and *point transition* on these normalized random measures, and construct a time dependent hierarchical model for dynamic topic modeling. By this, the dependencies are flexibly controlled between both jumps and atoms of the NRMs. All proofs and some extended theories are available in (Chen et al., 2012).

## 2. Normalized Random Measures

### 2.1. Background and Definitions

This background on random measures follows (James et al., 2009).

Let $(\mathbb{S}, \mathcal{S})$ be a measure space where $\mathcal{S}$ is the $\sigma$-algebra of $\mathbb{S}$. Let $\nu$ be a measure on it. A *Poisson process* on $\mathbb{S}$ is a random subset $\Pi \in \mathbb{S}$ such that if $N(A)$ is the number of points of $\Pi$ in the measurable set $A \subseteq \mathbb{S}$, then $N(A)$ is a Poisson random variable with mean $\nu(A)$, and $N(A_1), \cdots, N(A_n)$ are independent if $A_1, \cdots, A_n$ are disjoint.

Based on the definition, we define a complete random measure (CRM) on $(\mathbb{X}, \mathcal{B}(\mathbb{X}))$ to be a linear functional of the Poisson random measure $N(\cdot)$, with *mean measure* $\nu(\mathrm{d}t, \mathrm{d}x)$ defined on a product space $\mathbb{S} = R^+ \times \mathbb{X}$:

$$\tilde{\mu}(B) = \int_{\mathbb{R}^+ \times B} tN(\mathrm{d}t, \mathrm{d}x), \forall B \in \mathcal{B}(\mathbb{X}). \qquad (1)$$

Here $\nu(\mathrm{d}t, \mathrm{d}x)$ is called the *Lévy measure* of $\tilde{\mu}$.

It is worth noting that the CRM is usually written in the form $\tilde{\mu}(B) = \sum_{k=1}^{\infty} J_k \delta_{x_k}(B)$, where $J_1, J_2, \cdots > 0$ are called the *jumps* of the process, and $x_1, x_2, \cdots$ are a sequence of independent random variables drawn from a *base measurable space* $(\mathbb{X}, \mathcal{B}(\mathbb{X}))$[1]. A *normalized random measure* (NRM) on $(\mathbb{X}, \mathcal{B}(\mathbb{X}))$ is defined as $\mu = \frac{\tilde{\mu}}{\tilde{\mu}(\mathbb{X})}$. We always use $\mu$ to denote an NRM, and $\tilde{\mu}$ its unnormalized counterpart.

Taking different Lévy measures $\nu(\mathrm{d}t, \mathrm{d}x)$, we can obtain different NRMs, and the form we consider is described in Section 2.3. Here we consider the case $\nu(\mathrm{d}t, \mathrm{d}x) = M\rho_\eta(\mathrm{d}t)H(\mathrm{d}x)$, where $H(\mathrm{d}x)$ is the *base probability measure*, $M$ is the *total mass* acting as a concentration parameter, and $\eta$ is the set of other hyperparameters, depending on the specific NRM's. We use $\mathrm{NRM}(M, \eta, H)$ to denote the corresponding normalized random measure.

[1] $\mathcal{B}(\mathbb{X})$ means the $\sigma$-algebra of $\mathbb{X}$, we sometimes omit this and use $\mathbb{X}$ to denote the measurable space.

### 2.2. Slice sampling NRMs

We briefly introduce the ideas of slice sampling normalized random measures discussed in ("Slice 1" version, Griffin & Walker, 2011). It deals with the normalized random measure mixture of the type

$$\mu(\cdot) = \sum_{k=1}^{\infty} r_k \delta_{\theta_k}(\cdot),\ \theta_{s_i} \sim \mu(\cdot),\ x_i \sim g_0(\cdot|\theta_{s_i}) \qquad (2)$$

where $r_k = J_k / \sum_{l=1}^{\infty} J_l$, $\theta_k$'s are the component of the mixture model drawn *i.i.d.* from a parameter space $H(\cdot)$, $s_i$ denotes the component that $x_i$ belongs to, and $g_0(\cdot|\theta_k)$ is the density function to generate data from component $k$. Given the observations $\vec{x}$, a slice latent variable $u_i$ is introduced for each $x_i$ so that it only considers those components whose jump sizes $J_k$'s are larger than the corresponding $u_i$'s. Furthermore, an auxiliary variable $v$ is introduced to decouple each individual jump $J_k$ and their infinite sum of the jumps $\sum_{l=1}^{\infty} J_l$ appeared in the denominators of $r_k$'s. It is shown in (Griffin & Walker, 2011) that the posterior of the infinite mixture model (2) with the above auxiliary variables is proportional to

$$P_\mu(\vec{\theta}, J_1, \cdots, J_K, K, \vec{u}, L, \vec{s}, v | \vec{x}, H, \rho_\eta) \propto$$
$$\exp\left\{-v \sum_{k=1}^{K} J_k\right\} \exp\left\{-M \int_0^L (1 - \exp\{-vt\}) \rho_\eta(t)\mathrm{d}t\right\}$$
$$v^{N-1} p(J_1, \cdots, J_K) \prod_{k=1}^{K} h(\theta_k) \prod_{i=1}^{N} 1(J_{s_i} > u_i) g_0(x_i|\theta_{s_i}), \qquad (3)$$

where $1(a)$ is an indicator function returning 1 if $a$ is true and 0 otherwise, $h(\cdot)$ is the density of $H(\cdot)$, $L = \min\{\vec{u}\}$, and $p(J_1, \cdots, J_K) = \prod_{k=1}^{K} \frac{\rho_\eta(J_k)}{\int_L^\infty \rho_\eta(t)\mathrm{d}t}$ is the distribution for the jumps which are larger than $L$ derived from the underlying Poisson process. Sampling for this mixture model iteratively cycles over $\{\vec{\theta}, (J_1, \cdots, J_K), K, \vec{u}, \vec{s}, v\}$ based on (3). Please refer to (Section 1.3 Chen et al., 2012) for more details.

### 2.3. Normalized generalized Gamma processes

In this paper, we consider the normalized generalized Gamma processes. Generalized Gamma processes (Lijoi et al., 2007) (GGP) are random measures with the Lévy measure

$$\nu(\mathrm{d}t, \mathrm{d}x) = M \frac{e^{-bt}}{t^{1+a}} H(\mathrm{d}x), b > 0, 0 < a < 1. \qquad (4)$$

By normalizing the GGP, we obtain the normalized generalized Gamma process (NGG)[2]. One of the most familiar special cases is the *Dirichlet process*, which is a normalized Gamma process where $a \to 0$ and $b = 1$ and the concentration parameter appears as $M$.

Crucially, unlike the DP, the NGG can produce the *power-law* phenomenon.

**Proposition 1 ((Lijoi et al., 2007))** *Let $K_n$ be the number of components induced by the NGG with parameters $a$ an $b$ or the Dirichlet process with total mass $M$. Then for the NGG, $K_n/n^a \to S_{ab}$ almost surely, where $S_{ab}$ is a strictly positive random variable parameterized by $a$ and $b$. For the DP, $K_n/\log(n) \to M$.*

Therefore, in order to better analyze certain kinds of real data, we propose to use the NGG in place of the Dirichlet process. In the next section, we propose a dynamic topic model which extends two major advances of the Dirichlet process: the HDP (Teh et al., 2006) and the dependent Dirichlet process (Lin et al., 2010), to normalized random measures.

## 3. Dynamic topic modeling with dependent hierarchical NRMs

Our main interest is to construct a dynamic topic model that inherits *partial* exchangeability, meaning that the documents within each time frame are exchangeable, while between time frames they are not. To achieve this, it is crucial to model the dependency of the topics between different time frames. In particular, a topic can either inherit from the topics of earlier time frames with certain transformation, or be a completely new one which is "born" in the current time frame. The above idea can be modeled by a series of hierarchical NRMs, one per time frame. Between the time frames, these hierarchical NRMs depend on each other through three dependency operators: *superposition*, *subsampling* and *point transition*, which will be defined below. The corresponding graphical model is shown in Figure 1(left) and the generating process for the model is as follows:

- Generating independent NRMs $\mu_m$ for time frame $m = 1, \cdots, n$:

$$\mu_m | H, \eta_0 \sim \mathrm{NRM}(M_0, \eta_0, P_0) \tag{5}$$

  where $H(\cdot) = M_0 P_0(\cdot)$. $M_0$ is the total mass for $\mu_m$ and $P_0$ is the base distribution. In this paper, $P_0$ is the Dirichlet distribution, $\eta_0$ is the set of hyperparameters of the corresponding NRM, *e.g.*, in NGG, $\eta_0 = \{a, b\}$.

- Generating dependent NRMs $\mu'_m$ (from $\mu_m$ and $\mu'_{m-1}$), for time frame $m > 1$:

$$\mu'_m = T(S^q(\mu'_{m-1})) \oplus \mu_m \ . \tag{6}$$

  where the three dependency operators *superposition* ($\oplus$), *subsampling* ($S^q(\cdot)$) with acceptance rate $q$, and *point transition* ($T(\cdot)$) are generalized from those of Dirichlet process (Lin et al., 2010). We will discuss them in more details in the following subsection.

- Generating hierarchical NRM mixtures ($\mu_{mj}$, $\theta_{mji}$, $x_{mji}$) for time frame $m = 1, \cdots, n$, document $j = 1, \cdots, N_m$, word $i = 1, \cdots, W_{mj}$:

$$\mu_{mj} = \mathrm{NRM}(M_m, \eta_m, \mu'_m), \tag{7}$$
$$\theta_{mji} | \mu_{mj} \sim \mu_{mj}, \qquad x_{mji} | \theta_{mji} \sim g_0(\cdot | \theta_{mji})$$

  where $M_m$ is the total mass for $\mu_{mj}$, $g_0(\cdot|\theta_{mji})$ denotes the density function to generate data $x_{mji}$ from atom $\theta_{mji}$.

### 3.1. The three dependency operators

Adapting from the dependent Dirichlet process (Lin et al., 2010), the three dependency operators for the NRMs are defined as follows.

**Superposition of normalized random measures** Given $n$ independent NRMs $\mu_1, \cdots, \mu_n$ on $\mathbb{X}$, the superposition ($\oplus$) is:

$$\mu_1 \oplus \mu_2 \oplus \cdots \oplus \mu_n := c_1\mu_1 + c_2\mu_2 + \cdots + c_n\mu_n \ .$$

where the weights $c_m = \frac{\tilde{\mu}_m(\mathbb{X})}{\sum_j \tilde{\mu}_j(\mathbb{X})}$ and $\tilde{\mu}_m$ is the unnormalized version of $\mu_m$.

**Subsampling of normalized random measures** Given a NRM $\mu = \sum_{k=1}^{\infty} r_k \delta_{\theta_k}$ on $\mathbb{X}$, and a Bernoulli parameter $q \in [0, 1]$, the subsampling of $\mu$, is defined as

$$S^q(\mu) := \sum_{k: z_k = 1} \frac{r_k}{\sum_j z_j r_j} \delta_{\theta_k}, \tag{8}$$

where $z_k \sim \mathrm{Bernoulli}(q)$ are Bernoulli random variables with acceptance rate $q$.

**Point transition of normalized random measures** Given a NRM $\mu = \sum_{k=1}^{\infty} r_k \delta_{\theta_k}$ on $\mathbb{X}$, the point transition of $\mu$, is to draw atoms $\theta'_k$ from a transformed base measure to yield a new NRM as $T(\mu) := \sum_{k=1}^{\infty} r_k \delta_{\theta'_k}$.

[2]In NGG, $b$ can be absolved into $M$, thus we usually set $b = 1$, see (Chen et al., 2012) for detail.

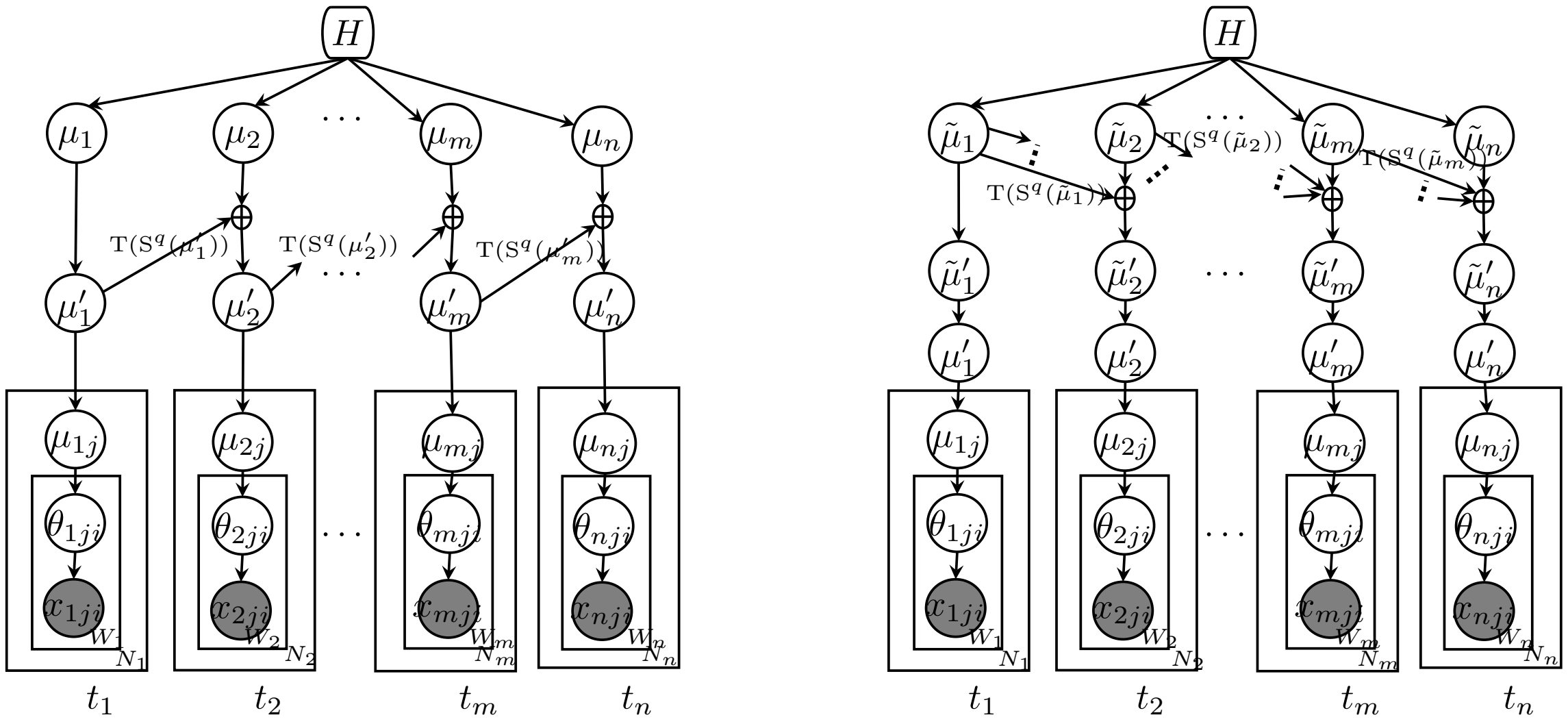

*Figure 1.* The time dependent topic model. The left plot corresponds to directly manipulating on normalized random measures (9), the right one corresponds to manipulating on unnormalized random measures (10). T: Point transition; $S^q$: Subsampling with acceptance rate $q$; $\oplus$: Superposition. Here $m = n - 1$ in the figures.

Point transitions can be done in different ways with different transition kernels $T(\cdot)$. In this paper, following (Lin et al., 2010), when inheriting from NRM $\mu$, we draw atoms $\theta'_k$ from the base measure as $\mu$ conditioned on its current statistics. Other ways of constructing transition kernels are left for further research.

### 3.2. Properties of the dependency operators

The three dependency operators on the NRMs inherit some of the nice properties from the underlying Poisson process. It not only enables quantitatively controlling dependencies introduced after and before the operations, as is shown in (Section 4 Chen et al., 2012), but also maintains a nice equivalence relation between the NRM's and the corresponding CRM's. In the following theorem, we will use $\tilde{\oplus}$, $\tilde{S}^q(\cdot)$ and $\tilde{T}(\cdot)$ to denote the three operations on their corresponding CRM's[3].

**Theorem 2** *The following time dependent random measures (9) and (10) are equivalent:*

- *Manipulate the normalized random measures:*

$$\mu'_m \sim T(S^q(\mu'_{m-1})) \oplus \mu_m, m > 1. \tag{9}$$

- *Manipulate the completely random measures:*

$$\tilde{\mu}'_m \sim \tilde{T}(\tilde{S}^q(\tilde{\mu}'_{m-1})) \oplus \tilde{\mu}_m, m > 1.$$
$$\mu'_m = \frac{\tilde{\mu}'_m}{\tilde{\mu}'_m(\mathbb{X})}, \tag{10}$$

[3]The definitions of $\tilde{\oplus}$, $\tilde{S}^q(\cdot)$ and $\tilde{T}(\cdot)$ are similar to the NRMs', see (Section 1.5.1 Chen et al., 2012) for details.

*Furthermore, the resulting NRMs $\mu'_m$'s give the following:*

$$\mu'_m = \sum_{j=1}^{m} \frac{\left(q^{m-j}\tilde{\mu}_j\right)(\mathbb{X})}{\sum_{j'=1}^{m}\left(q^{m-j'}\tilde{\mu}_{j'}\right)(\mathbb{X})} T^{m-j}(\mu_j), m > 1$$

*where $q^{m-j}\tilde{\mu}$ is the random measure with Lévy measure $q^{m-j}\nu(\mathrm{d}t, \mathrm{d}x)$ ($\nu(\mathrm{d}t, \mathrm{d}x)$ is the Lévy measure of $\tilde{\mu}$). $T^{m-j}(\mu)$ denotes point transition on $\mu$ for $(m-j)$ times .*

### 3.3. Reformulation of the proposed model

Theorem 2 in the last section allows us to first take *superposition*, *subsampling*, and *point transition* on the completely random measures $\tilde{\mu}_g$'s and then do the normalization. Therefore, we make use of Theorem 2 to obtain the dynamic topic model in Figure 1(right) by expanding the recusive formula in (10), which is equivalent to the left one.

The generating process of the new model is:

- Generating independent CRM's $\tilde{\mu}_m$ for time frame $m = 1, \cdots, n$, following (1).
- Generating $\mu'_m$ for time frame $m > 1$, following (10).
- Generating hierarchical NRM mixtures ($\mu_{mj}$, $\theta_{mji}$, $x_{mji}$) following (7).

The reason for this reformulation is because the inference on the model in Figure 1(left) appears to be infeasible. In general, the posterior of an NRM introduce complex dependencies between jumps, thus sampling is unclear after taking the three dependency operators.

On the other hand, the model in Figure 1(right) is more amenable to computation because the NRMs and the three operators are decoupled. It allows us to first generate the dependent CRM's, then use the slice sampler introduced in Section 2.2 to sample the posterior of the corresponding NRMs. From now on, we will focus on the model in Figure 1(right). In the next section, we discuss its sampling procedure.

## 4. Sampling

To introduce our sampling method we use the familiar Chinese restaurant metaphor (*e.g.* (Teh et al., 2006)) to explain key statistics. In this model customers for the variable $\mu_{mj}$ correspond to words in a document, restaurants to documents, and dishes to topics. In time frame $m$,

- $x_{mji}$: the customer $i$ in the $j$th restaurant.
- $s_{mji}$: the dish that $x_{mji}$ is eating.
- $n_{mjk}$: $n_{mjk} = \sum_i \delta_{s_{mji=k}}$, the number of customers in $\mu_{mj}$ eating dish $k$.
- $t_{mjr}$: the table $r$ in the $j$th restaurant.
- $\psi_{mjr}$: the dish that the table $t_{mjr}$ is serving.
- $n'_{mk}$: $n'_{mk} = \sum_j \sum_r \delta_{\psi_{mjr}=k}$, the number of customers[4] in $\mu'_m$ eating dish $k$.
- $\tilde{n}'_{mk}$: $\tilde{n}'_{mk} = n'_{mk}$, the number of customers in $\tilde{\mu}'_m$ eating dish $k$.
- $\tilde{n}_{mk}$: $\tilde{n}_{mk} = \sum_{m' \geq m} \tilde{n}'_{m'k}$, the number of customers in $\tilde{\mu}_m$ eating dish $k$.

We will do the sampling by marginalizing out $\mu_{mj}$'s. As it turns out, the remaining random variables that require sampling are $s_{mji}$, $n'_{mk}$, as well as

$$\tilde{\mu}_m = \sum_k J_{mk}\delta_{\theta_k}, \qquad \tilde{\mu}'_m = \sum_k J'_{mk}\delta_{\theta_k}$$

Note the $t_{mjr}$ and $\psi_{mjr}$ are not sampled as we sample the $n'_{mk}$ directly. Thus our sampler deals with the following latent statistics and variables: $s_{mji}$, $n'_{mk}$, $J_{mk}$, $J'_{mk}$ and some auxiliary variables are sampled to support these.

[4]the customers in $\mu'_m$ corresponds to the tables in $\mu_{mj}$. For convenient, we also regard a CRM as a restaurant.

**Sampling $J_{mk}$.** Given $\tilde{n}_{mk}$, we use the slice sampler introduced in (Griffin & Walker, 2011) to sample these jumps, with the posterior given in (3). Note that the mass $M_m$'s are also sampled, see (Sec.1.3 Chen et al., 2012). The resulting $\{J_{mk}\}$ are those jumps that exceed a threshold defined in the slice sampler, thus the number of jumps is finite.

**Sampling $J'_{mk}$.** $J'_{mk}$ is obtained by subsampling of $\{J_{m'k}\}_{m' \leq m}$[5]. By using a Bernoulli variable $z_{mk}$,

$$J'_{mk} = \begin{cases} J_{m'k} & \text{if } z_{mk} = 1 \\ 0 & \text{if } z_{mk} = 0. \end{cases}$$

We compute the posterior $p(z_{mk} = 1 | \tilde{\mu}_m, \{\tilde{n}'_{mk}\})$ to decide whether to inherit this jump to $\tilde{\mu}'_m$ or not. These posteriors are given in (Corollary 3 Chen et al., 2012). In practice, we found it mixes faster if we integrate out $z_{mk}$'s. (Lemma 9 Chen et al., 2012) shows that $q$-subsampling of a CRM with Lévy measure $\nu(\cdot)$ results in another CRM with Lévy measure $q\nu(\cdot)$, thus the jump sizes in the resultant CRM are scaled by $q$, meaning that $J'_{mk} = q^{m-m'} J_{m'k}$.

After the sampling of $\{J'_{mk}\}$, we normalize it and obtain the NRM $\mu'_m$, $\mu'_m = \sum_k r_{mk}\delta_{\theta_k}$ where $r_{mk} = J'_{mk} / \sum_{k'} J'_{mk'}$

**Sampling $s_{mji}$, $n'_{mk}$.** The following procedures are similar to sampling an HDP. The only difference is that $\mu_{mj}$ and $\mu'_m$ are NRMs instead of DPs. The sampling method goes as follows:

- **Sampling $s_{mji}$:** We use a similar strategy as the *sampling by direct assignment algorithm* for the HDP (Teh et al., 2006), the conditional posterior of $s_{mji}$ is:

  $$p(s_{mji} = k | \cdot) \propto (\omega_k + \omega_0 M_m r_{mk}) g_0(x_{mji} | \theta_k)$$

  where $\omega_0$ and $\omega_k$ depend on the corresponding Lévy measure of $\mu_{mj}$ (see (Theorem 2 James et al., 2009)). When $\mu_{mj}$ is a DP, then $\omega_k \propto n_{mjk}$ and $\omega_0 \propto 1$. When $\mu_{mj}$ is a NGG, $\omega_k \propto n_{mjk} - a$ and $\omega_0 \propto a(b + v_{mj})^a$, where $v_{mj}$ is the introduced auxiliary variables which can be sampled by an adaptive-rejection sampler using the posterior given in (Proposition 1 James et al., 2009).
- **Sampling $n'_{mk}$:** Using the similar strategy as in (Teh et al., 2006), we sample $n'_{mk}$ by simulating the (generalized) Chinese Restaurant Process, following the prediction rule (the probabilities of generating a new table or sitting on existing tables) of $\mu_{mk}$ in (Proposition 2 James et al., 2009).

[5] Since all the atoms across $\{\tilde{\mu}_{m'}\}$ are unique, $J'_{mk}$ is inherited from only one of $\{J_{m'k}\}$.

## 5. Experiments

### 5.1. Power-law in the NGG

We first investigate the power-law phenomena in the NGG, we sample it using the scheme of (James et al., 2009) and compare it with the DP in Figure 2.

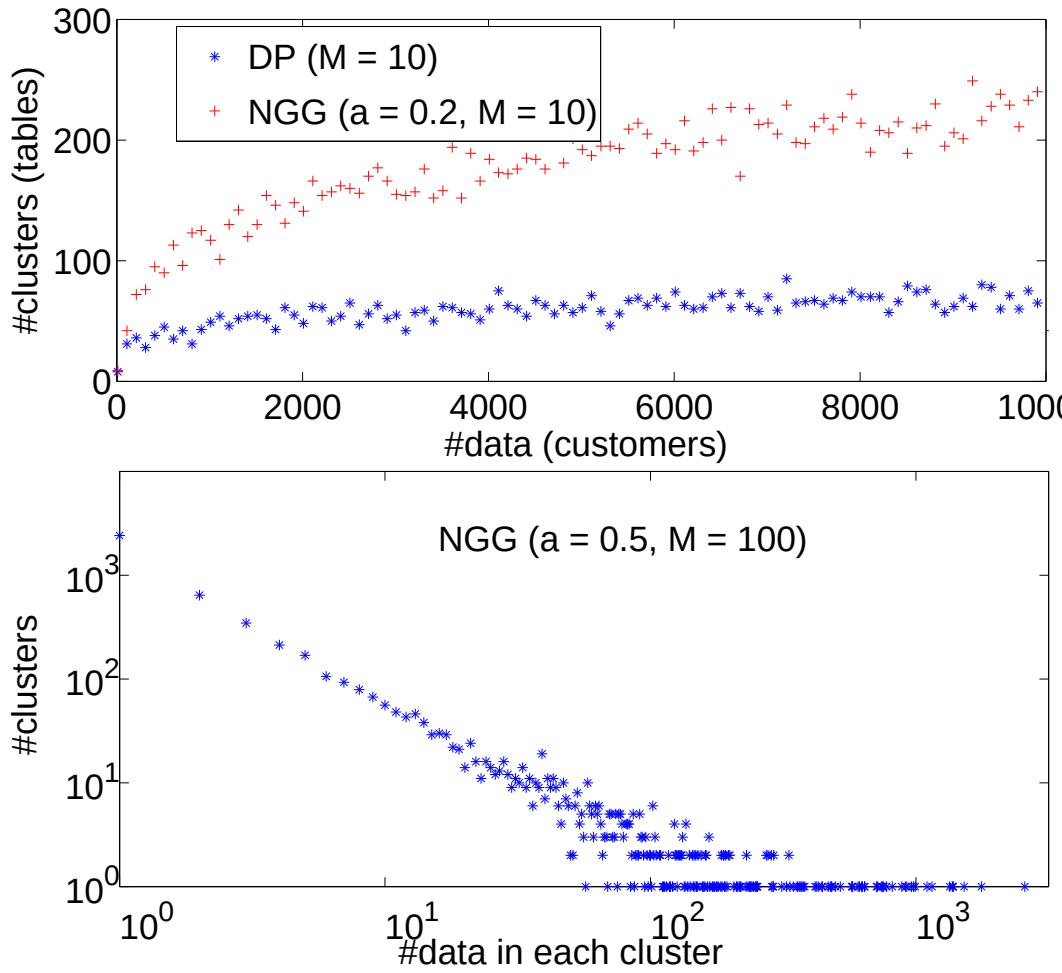

*Figure 2.* Power-law phenomena in NGG. The first plot shows the #data VS. #clusters, the second shows the size $s$ of each cluster VS. total number of clusters with size $s$.

### 5.2. Datasets

We tested our time dependent dynamic topic model on 9 datasets, removing stop-words and words appearing less than 5 times. ICML, JMLR, TPAMI are crawled from their websites and the abstracts are parsed. The preprocessed NIPS dataset is from (Globerson et al., 2007). The Person dataset is extracted from Reuters RCV1 using the query "person" under Lucene. The Twitter datasets are updates from three sports twitter accounts: `ESPN_FirstTake` (Twitter$_1$), `sportsguy33` (Twitter$_2$) and `SportsNation` (Twitter$_3$) obtained with the TweetStream API (http://pypi.python.org/pypi/tweetstream) to collect the last 3200 updates from each. The *Daily Kos* blogs (BDT) were pre-processed by (Yano et al., 2009). Statistics for the data sets are given in Table 1.

**Illustration:** Figure 3 gives an example of topic evolutions in the Twitter$_2$ dataset. We can clearly see that the three popular sports in the USA, *i.e.*, basketball, football and baseball, evolve reasonably with time. For example, MLB starts in April each year, showing a peak in baseball topic, and then slowly evolves with decreasing topic proportions. Also, in August one football topic is born, indicating a new season begins. Figure 4 gives an example of the word probability change in a single topic for the JMLR.

| dataset | vocab | docs | words | epochs |
|---|---|---|---|---|
| ICML | 2k | 765 | 44k | 2007–2011 |
| JMLR | 2.4k | 818 | 60k | 12 vols |
| TPAMI | 3k | 1108 | 91k | 2006–2011 |
| NIPS | 14k | 2483 | 3.28M | 1987-2003 |
| Person | 60k | 8616 | 1.55M | 08/96–08/97 |
| Twitter$_1$ | 6k | 3200 | 16k | 14 months |
| Twitter$_2$ | 6k | 3200 | 31k | 16 months |
| Twitter$_3$ | 6k | 3200 | 25k | 29 months |
| BDT | 8k | 2649 | 234k | 11/07–04/08 |

*Table 1.* Data statistics

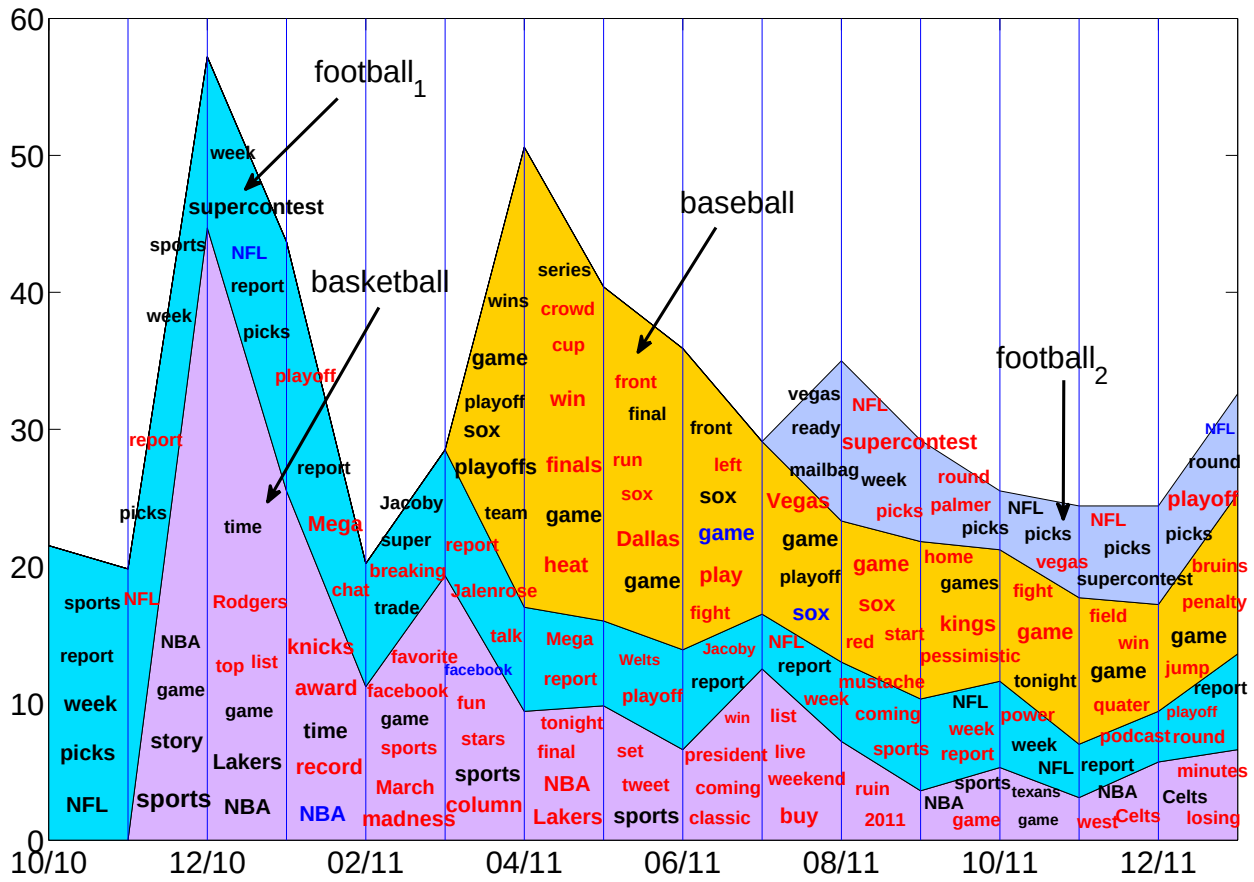

*Figure 3.* Topic evolution on Twitter. Words in red have increased, and blue decreased.

### 5.3. Quantitative Evaluations

**Comparisons** We first compare our model with two popular dynamic topic models where the author's own code was available for our use: (1) the dynamic topic model by Blei and Lafferty (Blei & Lafferty, 2006) and (2) the hierarchical Dirichlet process, where we used a three level HDP, with the middle level DP's representing the base topic distribution for the documents in a particular time. For fair comparison, similar to (Blei & Lafferty, 2006), we held out the data in previous time but used their statistics to help the training of the current time data, this is implemented in the HDP code by Teh. Furthermore, we also tested the proposed model without power-law, which is to use a DP instead of an NGG. We tested our model on the 9 datasets, for each dataset we used 80% for training and held out 20% for testing. The hyperparameters for DHNGG is set to $a = 0.2$ in this set of experiments with subsampling rate being 0.9, which is found to work well in practice. The topic-word distributions are symmetric Dirichlet with prior set to 0.3. Table 2 shows the test log-likelihoods for all these methods, which are calculated by first removing the test words from the topics and adding them back one by one and collecting the add-in probabilities as the testing likelihood (Teh et al., 2006). For all the methods we ran 2000 burn in iterations, followed by 200 iterations to collect samples. The results are averages over these samples.

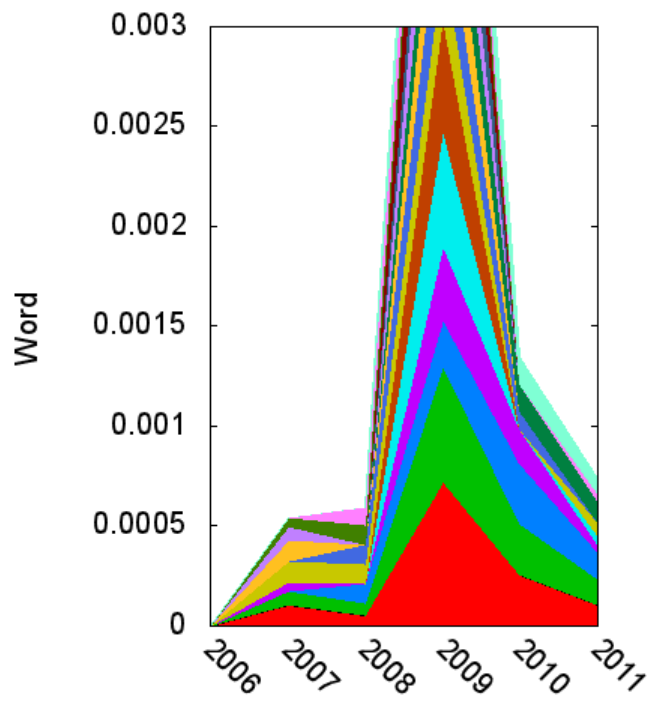

*Figure 4.* Topic evolution on JMLR. Shows a late developing topic on software, before during and after the start of MLOSS.org in 2008.

From Table 2 we see the proposed model *DHNGG* works best, with an improvement of 1%-3% in test log-likelihoods over the *HDP* model. In contrast the time dependent model *iDTM* of Ahmed & Xing (2010) only showed a 0.1% improvement over *HDP* on NIPS, implying the superiority of *DHNRM* over *iDTM*.

**Hyperparameter sensitivity** In NGG, there are hyperparameters $a$ and $b$, where $a$ controls the behavior of the power-law. In this section we study the influences of these two hyperparameters to the model. We varied $a$ among $(0.1, 0.2, 0.3, 0.5, 0.7, 0.9)$ while fixed the subsampling rate to 0.9 in this experiment. We run these settings on all these datasets, the training likelihoods are shown in Figure 5. From these results we consider $a = 0.2$ to be a good choice in practice.

**Influence of the subsampling rate** One of the distinct features of our model compared to other time dependent topic models is that the dependency comes partially from subsampling the previous time random measures, thus it is interesting to study the impact of subsampling rates to this model. In this experiment, we fixed $a = 0.2$, and varied the subsampling rate $q$ among $(0.1, 0.2, 0.3, 0.5, 0.7, 0.9, 1.0)$. The results are shown in Figure 6. From Figure 6, it is interesting to see that on the academic datasets, *e.g.*, ICML,JMLR, the best results are achieved when $q$ is approximately equal to 1; these datasets have higher correlations. While for the Twitter datasets, the best results are achieved when $q$ is equal to $0.5 \sim 0.7$, indicating that people tend to discuss more changing topics in these datasets.

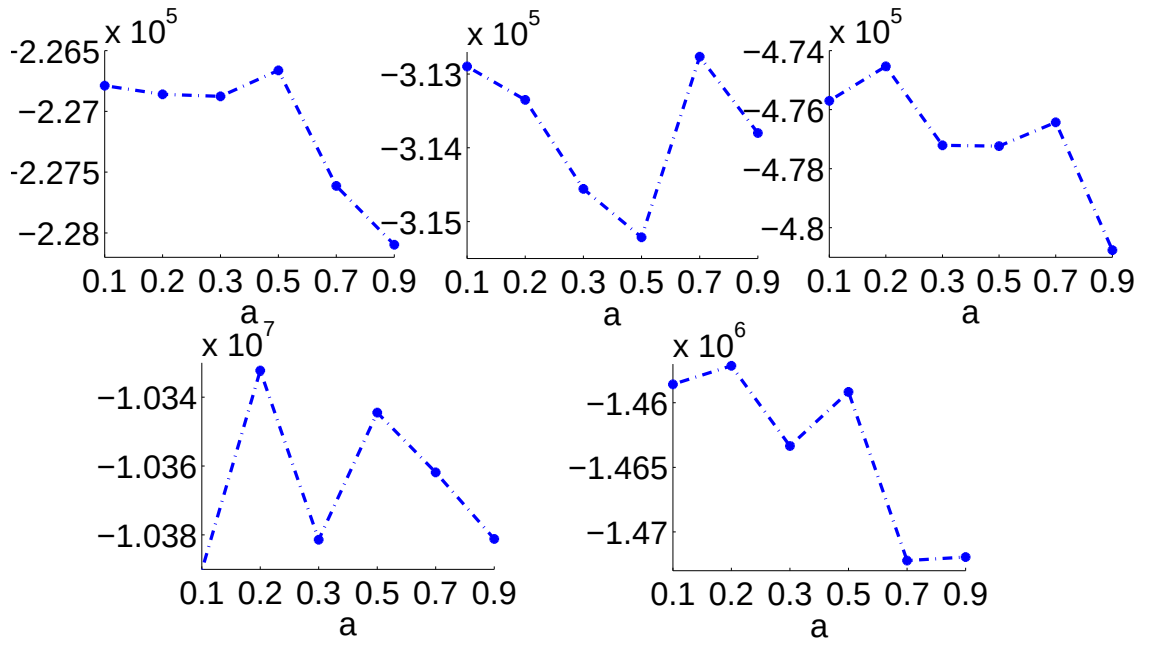

*Figure 5.* Training log-likelihoods influenced by hyperparameters $a$. From left to right (top-down) are the results on ICML, JMLR, TPAMI, Person and BDT.

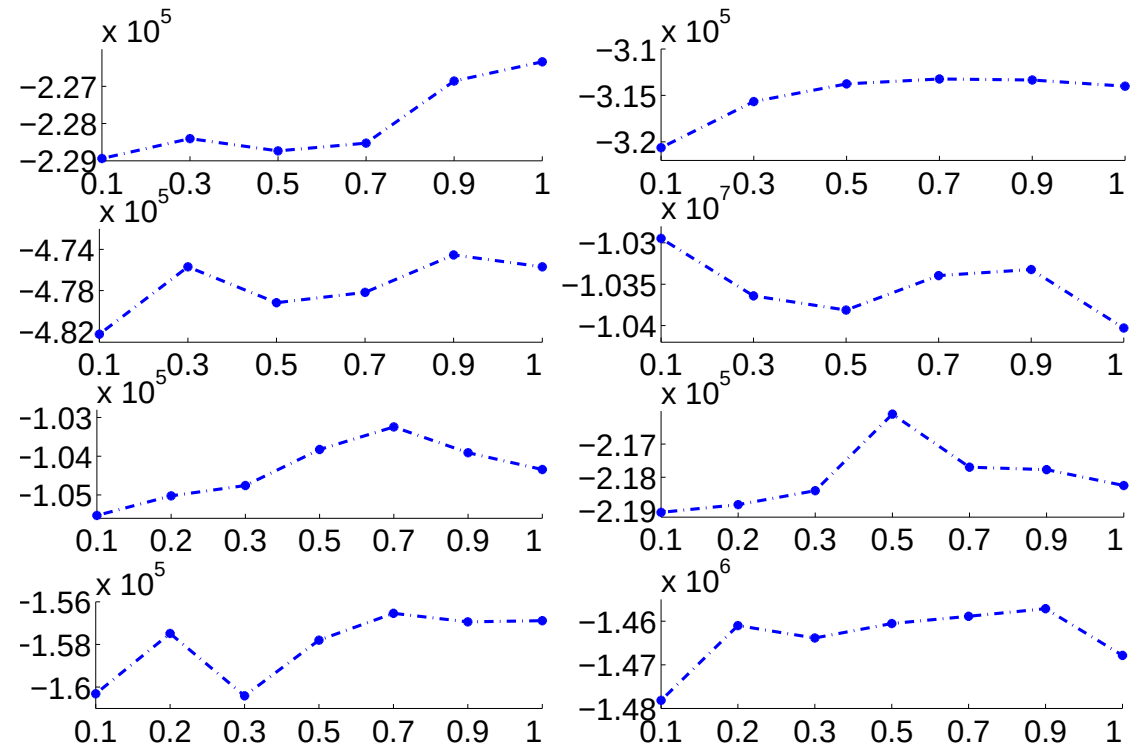

*Figure 6.* Training log-likelihoods influenced by the subsampling rate $q(\cdot)$. The $x$-axes represent $q$, the $y$-axes represent training log-likelihoods. From top-down, left to right are the results on ICML, JMLR, TPAMI, Person, $\text{Twitter}_1$, $\text{Twitter}_2$, $\text{Twitter}_3$ and BDT datasets, respectively.

## 6. Conclusion

We proposed dependent hierarchical normalized random measures. Specifically, we extend the three dependency operations for the Dirichlet process to normalized random measures and show how dependent models on NRMs can be implemented via dependent models on the underlying Poisson processes. Then we applied our model to dynamic topic modeling. Experimental results on different kinds of datasets demonstrate the superior performance of our model over existing models such as DTM, HDP and iDTM.

## Acknowledgments

We thank the reviewers for their valuable comments and Pinar Yanardag for collecting the Twitter data. NICTA

*Table 2.* Test log-likelihood on 9 datasets. *DHNGG*: dependent hierarchical normalized generalized Gamma processes, *DHDP*: dependent hierarchical Dirichlet processes, *HDP*: hierarchical Dirichlet processes, *DTM:* dynamic topic model (we set $K = \{10, 30, 50, 70\}$ and choose the best results).

| Datasets | ICML | JMLR | TPAMI | NIPS | Person |
|---|---|---|---|---|---|
| *DHNGG* | **-5.3123e+04** | **-7.3318e+04** | **-1.1841e+05** | **-4.1866e+06** | **-2.4718e+06** |
| *DHDP* | -5.3366e+04 | -7.3661e+04 | -1.2006e+05 | -4.4055e+06 | -2.4763e+06 |
| *HDP* | -5.4793e+04 | -7.7442e+04 | -1.2363e+05 | -4.4122e+06 | -2.6125e+06 |
| *DTM* | -6.2982e+04 | -8.7226e+04 | -1.4021e+05 | -5.1590e+06 | -2.9023e+06 |
| Datasets | $\text{Twitter}_1$ | $\text{Twitter}_2$ | $\text{Twitter}_3$ | BDT | |
| *DHNGG* | **-1.0391e+05** | **-2.1777e+05** | **-1.5694e+05** | **-3.3909e+05** | |
| *DHDP* | -1.0711e+05 | -2.2090e+05 | -1.5847e+05 | -3.4048e+05 | |
| *HDP* | -1.0752e+05 | -2.1903e+05 | -1.6016e+05 | -3.4833e+05 | |
| *DTM* | -1.2130e+05 | -2.6264e+05 | -1.9929e+05 | -3.9316e+05 | |

is funded by the Australian Government as represented by the Department of Broadband, Communications and the Digital Economy and the Australian Research Council through the ICT Center of Excellence program.

## References

Ahmed, A. and Xing, E.P. Timeline: A dynamic hierarchical Dirichlet process model for recovering birth/death and evolution of topics in text stream. In *UAI*, pp. 411–418, 2010.

Bartlett, N., Pfau, D., and Wood, F. Forgetting counts: constant memory inference for a dependent hierarchical Pitman-Yor process. In *ICML '10*. 2010.

Blei, D. and Lafferty, J. Dynamic topic models. In *ICML '06*. 2006.

Chen, C., Buntine, W., and Ding, N. Theory of dependent hierarchical normalized random measures. Technical Report arXiv:1205.4159, ANU and NICTA, Australia, May 2012. URL http://arxiv.org/abs/1205.4159.

Globerson, A., Chechik, G., Pereira, F., and Tishby, N. Euclidean Embedding of Co-occurrence Data. *JMLR*, 8:2265–2295, 2007.

Griffin, J.E. and Walker, S.G. Posterior simulation of normalized random measure mixtures. *J. Comput. Graph. Stat.*, 20(1):241–259, 2011.

James, L.F., Lijoi, A., and Prünster, I. Conjugacy as a distinctive feature of the Dirichlet process. *Scand. J. Stat.*, 33:105–120, 2006.

James, L.F., Lijoi, A., and Prünster, I. Posterior analysis for normalized random measures with independent increments. *Scand. J. Stat.*, 36:76–97, 2009.

Li, L., Zhou, M., Sapiro, G., and Carin, L. On the integration of topic modeling and dictionary learning. In *ICML*. 2011.

Lijoi, A., Mena, R.H., and Prünster, I. Controlling the reinforcement in Bayesian non-parametric mixture models. *J. R. Stat. Soc. Ser. B*, 69(4):715–740, 2007.

Lin, D., Grimson, E., and Fisher, J. Construction of dependent Dirichlet processes based on Poisson processes. In *NIPS*. 2010.

MacEachern, S. N. Dependent nonparametric processes. In *Proc. of the SBSS*. 1999.

Ren, L., Dunson, D.B., and Carin, L. The dynamic hierarchical Dirichlet process. In *ICML*, pp. 824–831, 2008.

Rubin, T., Chambers, A., Smyth, P., and Steyvers, M. Statistical topic models for multi-label document classification. Technical Report arXiv:1107.2462v2, University of California, Irvine, ASA, Nov 2011.

Socher, R., Maas, A., and Manning, C.D. Spectral Chinese restaurant processes: Nonparametric clustering based on similarities. In *AISTATS*. 2011.

Sudderth, E. B. and Jordan, M. I. Shared segmentation of natural scenes using dependent Pitman-Yor processes. In *NIPS*. 2008.

Teh, Y.W. A hierarchical Bayesian language model based on Pitman-Yor processes. In *ACL*, pp. 985–992, 2006.

Teh, Y.W., Jordan, M.I., Beal, M.J., and Blei, D.M. Hierarchical Dirichlet processes. *Journal of the ASA*, 101(476):1566–1581, 2006.

Yano, T., Cohen, W., and Smith, N.A. Predicting response to political blog posts with topic models. In *Proc. of the NAACL-HLT*, 2009.

Zhang, J., Song, Y., Zhang, C., and Liu, S. Evolutionary hierarchical Dirichlet processes for multiple correlated time-varying corpora. In *KDD*. 2010.